\documentclass{article}

\PassOptionsToPackage{square, sort, numbers}{natbib}

\usepackage[preprint]{arxiv_based_on_neurips_2020}

\usepackage[utf8]{inputenc} 
\usepackage[T1]{fontenc}    
\usepackage{hyperref}       
\usepackage{url}            
\usepackage{booktabs}       
\usepackage{amsfonts}       
\usepackage{nicefrac}       
\usepackage{microtype}      

\usepackage{epsfig}
\usepackage{graphicx}
\usepackage{amsmath}
\usepackage{amssymb}
\usepackage{mathtools}
\usepackage{subfig}
\usepackage{dsfont}
\usepackage{enumitem}
\usepackage{caption}
\usepackage{xspace}

\usepackage[symbol]{footmisc}

\DeclarePairedDelimiter\abs{\lvert}{\rvert}%
\usepackage[usenames, dvipsnames]{color} 
\definecolor{purple}{rgb}{0.5, 0.0, 0.5}
\definecolor{orange}{rgb}{1, 0.65, 0}
\definecolor{brown}{rgb}{0.64, 0.16, 0.16}
\iffalse 
  \newcommand{\freddy}[1]{\noindent}
  \newcommand{\corina}[1]{\noindent}
  \newcommand{\eric}[1]{\noindent}
  \newcommand{\todo}[1]{\noindent}
\else
  \newcommand{\freddy}[1]{\textcolor{purple}{\bf [FB: #1]}}
  \newcommand{\corina}[1]{\textcolor{orange}{\bf [CG: #1]}}
  \newcommand{\eric}[1]{\textcolor{brown}{\bf [EW: #1]}}
  \newcommand{\todo}[1]{\textcolor{red}{\bf [Todo: #1]}}
\fi

\newcommand{\missrate}[0]{MissRate\textsubscript{5, 2m}\xspace}
\newcommand{\ade}[1]{minADE\textsubscript{#1}\xspace}

\title{Motion Prediction using Trajectory Sets and Self-Driving Domain Knowledge}

\author{%
  Freddy A. Boulton\footnote{} \And Elena Corina Grigore \And Eric M. Wolff \\
  Motional \\
  \texttt{freddyalfonsoboulton@gmail.com, \{elena.corina.grigore, eric.wolff\}@motional.com}\\
}

\begin{document}

\footnotetext{\mbox{*}Work performed while at Motional.}

\maketitle

\begin{abstract}
Predicting the future motion of vehicles has been studied using various techniques, including stochastic policies, generative models, and regression.
Recent work has shown that classification over a trajectory set, which approximates possible motions, achieves state-of-the-art performance and avoids issues like mode collapse.
However, map information and the physical relationships between nearby trajectories is not fully exploited in this formulation.
We build on classification-based approaches to motion prediction by adding an auxiliary loss that penalizes off-road predictions.
This auxiliary loss can easily be \emph{pretrained} using only map information (e.g., off-road area), which significantly improves performance on small datasets.
We also investigate weighted cross-entropy losses to capture spatial-temporal relationships among trajectories.
Our final contribution is a detailed comparison of classification and ordinal regression on two public self-driving datasets.
\end{abstract}

\section{Introduction}
Self-driving cars must share the road with other vehicles, bicyclists, and pedestrians. 
To safely operate in this dynamic and uncertain environment, it is important to reason about the likely future motions of other road users.
Each road user has different goals (e.g., turn left, speed up, stop) and preferences (e.g., desired speed, braking profile), which influence his or her actions.
Thus, useful predictions of future motion must represent multiple possibilities and their associated likelihoods.

Motion prediction is fundamentally challenging due to the uncertainty in road user behavior.
Likely behaviors are constrained by both the road (e.g., following lanes, stopping for traffic lights) and the actions of others (e.g., slowing for a car ahead, stopping for a pedestrian crossing the road).

State-of-the-art motion prediction models learn features over a combined representation of the map and the recent history of other road users~\cite{cui2019multimodal,chai2019multipath,hong2019zoox}.
These methods use a variety of multimodal output representations, including stochastic policies~\cite{rhinehart2018r2p2}, occupancy maps~\cite{hong2019zoox}, and regression~\cite{cui2019multimodal,chai2019multipath}.
Recent work~\cite{phan2019cover-net} has shown that classification over a trajectory set, which approximates possible motions, achieves state-of-the-art performance and avoids issues like mode collapse.
However, that work did not use map data and trajectory set geometry in the loss, and had a limited comparison with ordinal regression.
This paper addresses those limitations.

The classification loss used in~\cite{phan2019cover-net} does not explicitly penalize predictions that go off the road.
To use the prior knowledge that cars typically drive on the road, we simplify and adapt the off-road loss from~\cite{niedoba2019uber-off-road} to classification over trajectory sets.
Our formulation allows \emph{pretraining} a model using only map data, which significantly improves performance for small datasets.

The trajectory sets introduced for motion prediction in~\cite{phan2019cover-net} have rich spatial-temporal structure.
However, the cross-entropy loss they used does not exploit the fact that classes correspond to physical trajectories.
We investigate weighted cross-entropy losses that determine how much ``near-misses'' in the trajectory set are penalized.
Surprisingly, we find that a variety of weighted losses do not improve on the original formulation.
We analyze these results in the context of prediction diversity.

The classification-based approach of~\cite{phan2019cover-net} is closely related to the ordinal regression approach of~\cite{chai2019multipath}, which computes residuals from a set of ``anchor'' trajectories that are analogous to a trajectory set.
We perform a detailed comparison of these approaches on two public datasets.
Interestingly, we find that performance of ordinal regression can benefit from using an order of magnitude more ``anchor'' trajectories than previously reported~\cite{chai2019multipath}.

Our main contributions on multimodal, probabilistic motion prediction are summarized as follows:
\begin{itemize}[nosep]
    \item use an off-road loss with pretraining to help learn domain knowledge;
    \item explore weighted cross-entropy losses to capture spatial relationships within a trajectory set;
    \item carefully compare classification vs ordinal regression on nuScenes~\cite{nuscenes2019} and Argoverse~\cite{chang2019argoverse}.
\end{itemize}

\section{Related Work}
\label{sec:related-work}
State-of-the-art motion prediction algorithms now typically use CNNs to learn appropriate features from a birds-eye-view rendering of the scene (map and road users).
Other road users are represented as sensor data~\cite{luo2018fast-and-furious,casas18intent-net} or the output of a tracking and fusion system~\cite{cui2019multimodal,hong2019zoox}.
Recent work~\cite{casas2019spaGNN,gao2018vector-net} has also explored using graph neural networks (GNNs) to encode interactions.

Complementing the input representations described above, various approaches have been used to represent the possible future motions.
Generative models encode choice over multiple actions via sampling latent variables.
Examples include stochastic policies~\cite{kitani2012activity-forecasting,rhinehart2018r2p2,rhinehart2019precog,tang2019mfp}, CVAEs~\cite{hong2019zoox, lee2017desire, bhattacharyya2018best-of-many, ivanovic2018multimodal-human} and GANs~\cite{sadeghian2019sophie-gan, gupta2018social-gan, zhao2019MultiAgentTF}.
These approaches require multiple samples or policy rollouts at inference.

Regression models either predict a single future trajectory~\cite{luo2018fast-and-furious, casas18intent-net, djuric2018shortterm, alahi2016social-lstm}, or a distribution over multiple trajectories~\cite{ cui2019multimodal, hong2019zoox, deo2018conv-pool-vehicles}.
The former unrealistically average over behaviors in many driving situations, while the latter can suffer from mode collapse.
In~\cite{chai2019multipath}, the authors propose a hybrid approach using ordinal regression.
This method regresses to residuals from a set of pre-defined anchors, much like in object detection.
This technique helps mitigate mode collapse.

CoverNet~\cite{phan2019cover-net} frames the problem as classification over a trajectory set, which approximates all possible motions. 
Our work further explores this formulation through losses that use the map and trajectory set geometry, as well as detailed experimental comparison to ordinal regression.

\subsection{Use of Domain Knowledge}
Both dynamic constraints and ``rules-of-the-road'' place strong priors on likely motions.
Dynamic constraints were explicitly enforced via trajectory sets in~\cite{phan2019cover-net} and a kinematic layer in~\cite{cui2019deep-kinematic}.

Prior work using map-based losses includes the use of an approximate prior that captures off-road areas as part of a symmetric KL loss~\cite{rhinehart2018r2p2}, and an off-road loss on future occupancy maps~\cite{bansal2019chauffeurnet}.
However, these loss formulations are not directly compatible with most trajectory prediction approaches that output point coordinates.
The most closely related works are~\cite{niedoba2019uber-off-road}, which applies an off-road loss to multimodal regression, and~\cite{casas2020importance}, which applies rewards for reachable lanes and the planned route of the autonomous vehicle.
Our formulation is simpler and allows pretraining using just the map.

We leverage domain knowledge by encoding the relationships among trajectories via a weighted cross-entropy loss, where the weight is a function of distance to the ground truth.
To our knowledge, this loss has not been explored in motion forecasting, likely due to the prior focus on generative and regression models.
This loss is typically used to mitigate class imbalance problems~\cite{chen2017learning}.

\subsection{Public datasets}
Until recently, public self-driving datasets suitable for motion prediction were either relatively small~\cite{geiger2012kitti} or for highway driving~\cite{colyar2007ngsim}.
Accordingly, many publications are evaluated only on private datasets~\cite{cui2019multimodal,zeng2019costmap,casas18intent-net,bansal2019chauffeurnet}.
We report results on two recently released self-driving datasets, nuScenes~\cite{nuscenes2019} and Argoverse~\cite{chang2019argoverse}, to help establish clear comparisons between state-of-the-art methods for motion forecasting on city roads.

\section{Preliminaries}
\label{sec:prelims}
We now set notation and give a brief, self-contained overview of CoverNet~\cite{phan2019cover-net}, which we will extend with modified loss functions in Section~\ref{sec:losses}. 
CoverNet uses the past states of all road users and a high-definition map to compute a distribution over a vehicle's possible future states.

\subsection{Notation}
\label{sec:notation}
We use the state outputs of an object detection and tracking system, and a high-definition map that includes lanes, drivable area, and other relevant information. 
Both are typically used by self-driving cars operating in urban environments and are also assumed in~\cite{phan2019cover-net,cui2019multimodal,chai2019multipath}.

We denote the set of road users at time $t$ by $\mathcal{I}_t$ and the state of object $i \in \mathcal{I}_t$ at time $t$ by $s^i_{t}$.
Let $s^{i}_{m:n} = \left[s^i_{m}, \ldots, s^i_{n}\right]$, where $m < n$ and $i \in \mathcal{I}_t$, denote the discrete-time trajectory of object $i$ at times $t = m, \ldots, n$.
Let $\mathcal{C} = \{ \bigcup_i s^{i}_{t-m:t}; \mathrm{map} \}$ denote the scene context over the past $m$ steps (i.e., partial history of all objects and the map). We are interested in predicting $s_{t:t+T}^i$ given $\mathcal{C}$, where $T$ is the prediction horizon.
\begin{figure*}[t]
\begin{center}
   \includegraphics[width=0.9\linewidth]{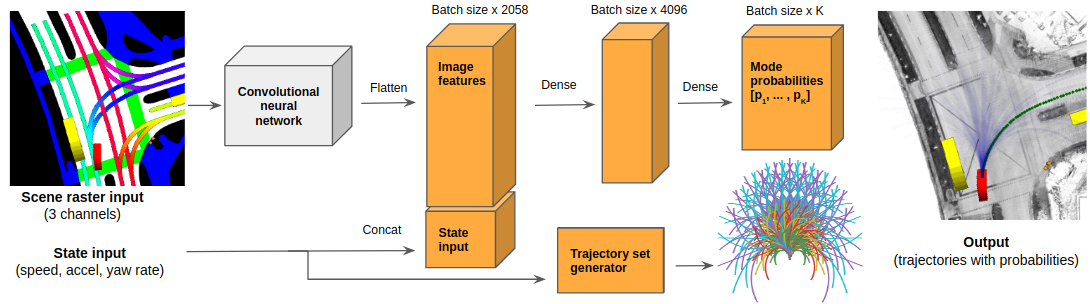}
\end{center}
   \caption{CoverNet overview from~\cite{phan2019cover-net}.}
\label{fig:network-architecture}
\vspace{0mm}
\end{figure*}
\subsection{CoverNet overview}
\label{sec:cover-net}

We briefly summarize CoverNet~\cite{phan2019cover-net}, which makes multimodal, probabilistic trajectory predictions for a vehicle of interest by classifying over a \emph{trajectory set}.
Figure~\ref{fig:network-architecture} overviews the model architecture.

\noindent\textbf{Input}
The input is a birds-eye-view raster image centered around the agent of interest that combines both map data and the past states of all objects.
The raster image is an RGB image that is aligned so that the agent's heading points up.
Map layers are drawn first, followed by vehicles and pedestrians.
The agent of interest, each map layer, and each object type are assigned different colors in the raster.
The sequence of past observations for each object is represented through fading object colors using linearly decreasing saturation (in HSV space) as a function of time.

\noindent\textbf{Output}
The output is a distribution over a \emph{trajectory set}, which approximates the vehicle's possible motions.  
The trajectory set is constructed from a larger dataset of representative trajectories using a coverage metric that helps construct a diverse set. 
The resulting set contains enough trajectories to ensure that the coverage metric for all trajectories in the dataset is bounded by $\epsilon$. 
Using a trajectory set is a reasonable approximation given the relatively short prediction horizons (3 to 6 seconds) and inherent uncertainty in agent behavior.
Let $\mathcal{K}$ be a trajectory set.
Then, the output dimension is equal to the number of modes, namely $\abs{\mathcal{K}}$.

The output layer applies the softmax function to convert network activations to a probability distribution.
The probability of the $k$-th trajectory is $p(s_{t:t+T}^k | \mathcal{C}) = \frac{\exp f_k(\mathcal{C})}{\sum_i\exp f_i(\mathcal{C})}$, where $f_i(\mathcal{C}) \in \mathbb{R}$ is the output of the network's penultimate layer.

\noindent\textbf{Training}
The model is trained as a multi-class classification problem using a standard cross-entropy loss.
For each example, the positive label is the element in the trajectory set closest to the ground truth, as measured by the minimum average of point-wise Euclidean distances.

\section{Losses}
\label{sec:losses}
We now introduce an off-road loss to partially encode ``rules-of-the-road,'' and a weighted cross-entropy classification loss to capture spatial relationships in a trajectory set.
\begin{figure}
    \centering
    \begin{minipage}{0.47\textwidth}
        \centering
        \includegraphics[scale=0.35]{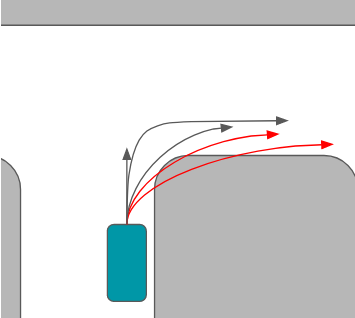}
        \caption{Visualization of on-road (black) and off-road (red) trajectories.}
        \label{fig:off-road-cartoon}
    \end{minipage}\hfill
    \begin{minipage}{0.47\textwidth}
        \centering
         \includegraphics[scale=0.35]{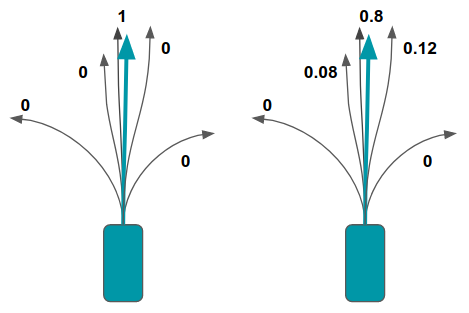}
        \caption{Visualization of the target distribution in the standard cross-entropy formulation (left), and the weighted cross-entropy loss (right).}
        \label{fig:weighted-ce-cartoon}
    \end{minipage}
\end{figure}
\subsection{Off-road loss}
\label{sec:off-road-loss}
Our intuition is that domain knowledge is helpful to exploit in the loss formulation.
In the self-driving domain, we expect most predictions to be inside of the driveable area of the road. 
For example, if we have a set of proposed trajectories like those in Figure~\ref{fig:off-road-cartoon}, we would like the loss to help the model learn
the black trajectories are more likely than the red ones.
With this in mind, we explore the effects of introducing an auxiliary loss that helps learn trajectories that are more compliant with respect to the driveable area.

Let $x_{t}^{i} \in \mathbb{R^{\abs{\mathcal{K}}}}$ be the activations of CoverNet's penultimate layer for agent $i$ at time $t$ and let $r_{t}^{i} \in \mathbb{R^{\abs{\mathcal{K}}}}$ be a binary vector whose value is $1$ at entry $k$ if trajectory $k$ in the set is entirely contained in the drivable area.
Let $\sigma(x)$ be the sigmoid function and let $\mathrm{bce}(\hat{y}, y) = (1 - y)\log(1 - \hat{y}) + y\log(\hat{y})$ be the binary cross-entropy loss.
We define the off-road loss as (with $[\cdot]$ denoting vector indexing):
\begin{equation}
\label{eq:off-road-loss}
\begin{aligned}
\Omega_{t}^{i} = \sum_{k=1}^{\abs{\mathcal{K}}} \mathrm{bce}(\sigma(x^t_i[k]), r^t_i[k]).
\end{aligned}
\end{equation}

We further introduce hyperparameter $\lambda \in \left[0, \infty\right)$ to encode the relative weight of the off-road loss versus the classification loss. The total loss for agent $i$ at time $t$ is then:
\begin{equation}
\label{eq:total-loss}
\begin{aligned}
\mathcal{L}_{t}^{i} = \mathcal{L}_{class}^{i} + \lambda \Omega_{t}^{i}.
\end{aligned}
\end{equation}

\subsection{Weighted cross-entropy loss}
\label{sec:modified-ce-loss}
The approach proposed in~\cite{phan2019cover-net} uses a cross-entropy loss where positive samples are determined by the element in the trajectory set closest to the ground truth.
This loss penalizes the second-closest trajectory just as much as the furthest, since it ignores the geometric structure of the trajectory set.

Our proposed modification is to create a probability distribution over all modes that are ``close enough'' to the ground truth.
So, instead of a delta distribution over the closest mode, there is also probability assigned to near misses (see Figure~\ref{fig:weighted-ce-cartoon}).

Let $D$ be a real-valued function that measures the distance between trajectories in trajectory set $\mathcal{K}$ and the ground truth for agent $i$ at time $t$ and let $d_{t}^{i} = D(\mathcal{K}, s_{t:t+T}^i) \in \mathbb{R^{\abs{\mathcal{K}}}}$.
We set a threshold that defines which trajectories are ``close enough'' to the ground truth and let $\mathcal{K}_{\mathrm{close}} \subset \mathcal{K}$ be the set of these trajectories.
We experiment with $D$ as both the max and mean of element-wise Euclidean distances, denoted by Max $\ell^2$ and Mean $\ell^2$, respectively.

We take the element-wise inverse of $d_{t}^{i}$ and set the value of trajectories not in $\mathcal{K}_{\mathrm{close}}$ to $0$.
We linearly normalize the entries of this vector so they sum to $1$ and use this vector as the target probability distribution with the standard cross-entropy loss and we denote it by $w_{t}^{i}$.
Finally, let $S$ be the softmax function and let $S(x_{t}^{i}) \in \mathbb{R^{\abs{\mathcal{K}}}}$ be the result of applying softmax to the model's penultimate layer. Our weighted cross entropy loss is defined as (with $[\cdot]$ denoting vector indexing):
\begin{equation}
\label{eq:modified-ce-loss}
\begin{aligned}
\mathcal{H}_{t}^{i} = \sum_{k=1}^{\abs{\mathcal{K}}} - w_{t}^{i}{}_k \textrm{log } S(x_{t}^{i})[k].
\end{aligned}
\end{equation}

\section{Experiments}
For our motion prediction experiments, the input representation and model architecture were fixed across all models and baselines.
We varied both the loss functions and output representations.

\subsection{Baselines}
\noindent\textbf{Physics oracle}.
The \emph{physics oracle}~\cite{phan2019cover-net} is the minimum average point-wise Euclidean distance over the following physics-based models: i) constant velocity and yaw, ii) constant velocity and yaw rate, iii) constant acceleration and yaw, and iv) constant acceleration and yaw rate.

\noindent\textbf{CoverNet~\cite{phan2019cover-net}}.
As described in Section~\ref{sec:cover-net}.
\begin{figure*}
    \centering
    \subfloat[][CoverNet, $\varepsilon = 2$, $\lambda = 0$]{%
        \includegraphics[width=0.3\textwidth]{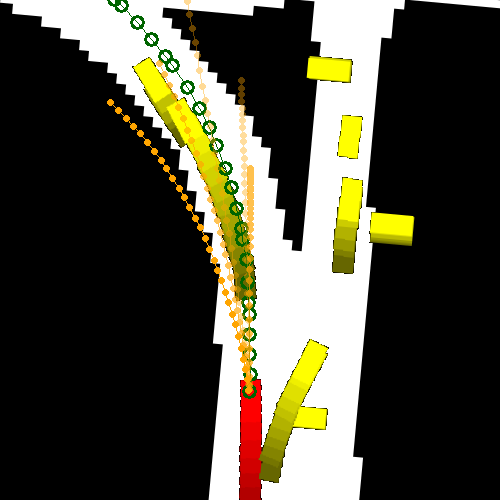}%
        }%
        \hspace{1mm}
    \subfloat[][CoverNet, $\varepsilon = 2$, $\lambda = 1$]{%
        \includegraphics[width=0.3\textwidth]{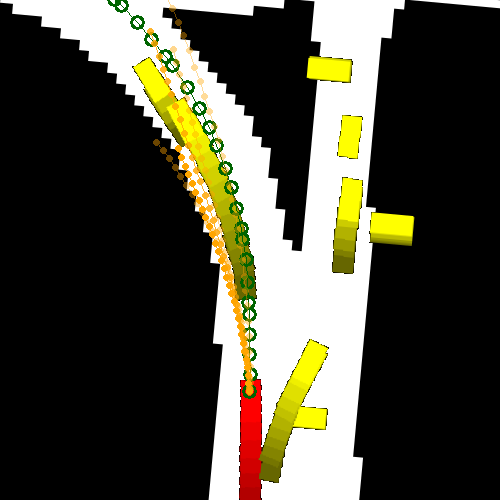}%
        }%
        \hspace{1mm}
    \subfloat[][CoverNet, $\varepsilon = 2$, $\lambda = 10$]{%
        \includegraphics[width=0.3\textwidth]{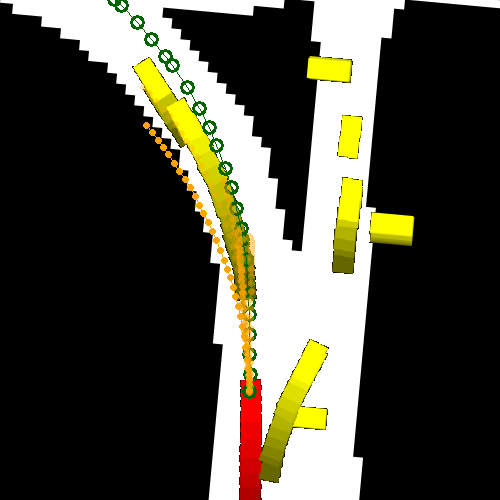}%
        }%
    \hfill%
    \caption{Examples of predicted trajectories on the same Argoverse scenario. They include a CoverNet model with $\varepsilon=2m$, in increasing order of off-road penalty. Objects in the world are rendered up to the current time. Note higher drivable area compliance as the penalty increases.}
    \label{fig:image-examples}
\end{figure*}

\noindent\textbf{Regression to anchor residuals}.
\label{sec:multi-path}
We compare our contributions to MultiPath (MP)~\cite{chai2019multipath}, a state-of-the-art multimodal regression model.
This model implements ordinal regression by first choosing among a fixed set of anchors (computed a priori) and then regressing to residuals from the chosen anchor.
This model predicts a fixed number of trajectories (modes) and their associated probabilities.
Each trajectory is then the sum of the corresponding anchor and predicted residual.
In our implementation, we use CoverNet trajectory sets as the fixed anchors.

\subsection{Implementation details}
Our implementation largely follows~\cite{cui2019multimodal} and~\cite{phan2019cover-net}.
See Figure~\ref{fig:network-architecture} for an overview.

The input raster is an RGB image of size ($H$, $W$, $3$).
The image is aligned so that the agent's heading faces up, and the agent is placed on pixel ($l$, $w$) as measured from the top-left of the image.
In our experiments, we use a resolution of $0.25$ meters per pixel and choose $l = 320$ and $w = 200$.
Thus, the model can “see” $80$ meters ahead, $20$ meters behind, and $25$ meters on each side of the agent.
Objects are rendered based on their oriented bounding box, which are imputed for Argoverse.

All models use a ResNet-50~\cite{he2015resnet} backbone with pretrained ImageNet~\cite{russakovsky2015image-net} weights~\cite{pytorch-models}.
We apply a global pooling layer to the ResNet \emph{conv5} feature map and concatenate the result with the agent's speed, acceleration, and yaw rate.
We then add a $4096$ dimensional fully connected layer.

Trajectory sets for each dataset were constructed as described in~\cite{phan2019cover-net}.

For the regression models, our outputs are of dimension $\abs{\mathcal{K}} \times (\abs{\vec{x}} \times N + 1)$, where $\abs{\mathcal{K}}$ represents the total number of predicted modes, $\abs{\vec{x}}$ represents the number of coordinates we are predicting per point, $N$ represents the number of points in our predictions, and the extra output per mode is the probability associated with each mode.
For our implementations, $N = H \times F$, where $H$ is the length of the prediction horizon, and $F$ is the sampling frequency.
We predict $(x, y)$ coordinates, so $|\vec{x}| = 2$.

We use a smooth $\ell^1$ loss for all regression baselines.

Models were trained for $20$ epochs on a $4$ GPU machine, using a batch size of $32$, and a learning rate schedule that starts at $1\mathrm{e-}3$ and applies a multiplicative factor of $0.9$ every epoch.
We found that these parameters gave good performance across all models.
The per-batch core time was \textasciitilde$300$ ms.

\subsection{Metrics}
\label{subsec:metrics}
We use commonly reported metrics to give insight into various facets of multimodal trajectory prediction.
In our experiments, we average the metrics described below over all instances.

For insight into trajectory prediction performance in scenarios where there are multiple plausible actions, we use the minimum average displacement error (ADE).
The \textrm{\ade{k}} is \( \min_{\hat{s} \in \mathcal{P}} \frac{1}{N}\sum_{\tau=t}^{t+N} || s_\tau - \hat{s}_\tau || \), where $\mathcal{P}$ is the set of $k$ most likely trajectories.

We use the notion of a \emph{miss rate} to simplify interpretation of whether or not a prediction was ``close enough.''
We define the $\textrm{MissRate}_{k, d}$ for a single instance (agent at a given time) as 1 if \(\min_{\hat{s} \in \mathcal{P}} \max_{\tau=t}^{t+N} || s_\tau - \hat{s}_\tau || \leq d \), and 0 otherwise.

To measure how well predicted trajectories satisfy the domain knowledge that cars drive on the road, we use Drivable Area Compliance (DAC)~\cite{chang2019argoverse}.
DAC is computed as $(n - n_{\mathrm{offroad}}) / n$, where $n$ is the total number of predictions and $n_{\mathrm{offroad}}$ is the number that are off the road.

\subsection{The impact of domain knowledge}
\label{subsec:argoverse}
We investigate the impact of the domain knowledge that we encoded with the losses in Section~\ref{sec:losses}.
All results in this section are evaluated on the Argoverse~\cite{chang2019argoverse} validation set.

\noindent\textbf{Argoverse}.
Argoverse~\cite{chang2019argoverse} is a large-scale motion prediction dataset recorded in Pittsburgh and Miami.
Object centroids (imperfectly estimated by a tracking system) are published at $10$ Hz.
The task is to predict the last $3$ seconds of each scene given its first $2$ seconds and the map.
We use the split for version $1.1$ of the dataset, which includes $205,942$ training scenarios.

\subsubsection{Off-road loss and pretraining}
\begin{figure}
    \centering
    \includegraphics[width=1.0\textwidth]{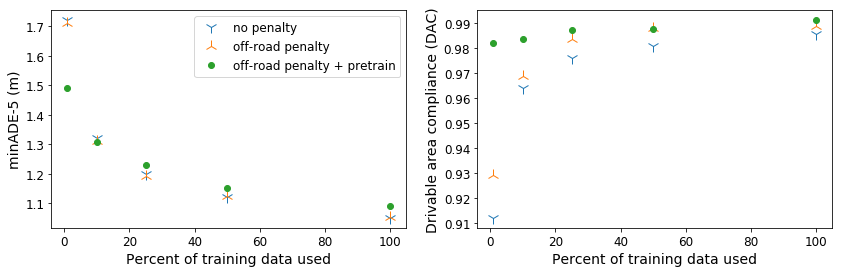}
    \caption{Effect of limited training data. The off-road penalty is $\lambda=1$.}
    \label{fig:pretrain-data-size}
\end{figure}

As all vehicles in Argoverse are on the drivable area, a perfect model should have drivable area compliance (DAC) of $1$.
Figure~\ref{fig:pretrain-data-size} explores how model performance varies with the amount of training data available.
Using an off-road penalty improves DAC in all cases.
We also investigated the effects of pretraining our model on map data only.
It is easy to create arbitrarily many training examples for the off-road loss with only map data.
To do this, we pretrained our models on the full map-only Argoverse train set.
We then loaded the pretrained models and continued training with the full information, varying the amount of data from the train set.
As Figure~\ref{fig:pretrain-data-size} suggests, pretraining gives a significant performance boost in data-limited regimes.

\begin{figure}
    \centering
    \begin{minipage}{0.46\textwidth}
        \begin{center}
        \includegraphics[width=\linewidth]{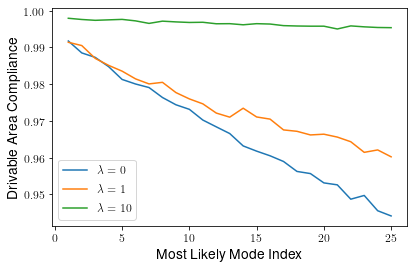}
        \end{center}
        \caption{DAC ordered by mode probability.}
        \label{fig:off-road-mode-prob}
    \end{minipage}\hfill
    \begin{minipage}{0.46\textwidth}
        \centering
        \begin{tabular}{l cc}
        \toprule
        Anchors & $\ell^1$ & $\ell^{inf}$ \\
        \midrule
        16   & 0.70 & 63.67 \\
        64   & 0.49 & 74.34 \\
        682  & 0.29 & 5.76 \\
        1571 & 0.18 & 2.93 \\
        \bottomrule
        \end{tabular}
        \caption{Ordinal regression residuals (m) on the Argoverse val set.}
        \label{tab:residuals}
    \end{minipage}
\end{figure}

Figure~\ref{fig:off-road-mode-prob} further explores the impact of off-road losses on DAC of less likely modes.
We see small but consistent improvements in DAC as we increase the off-road loss penalty.
Furthermore, the relative improvements increase with less likely modes.
We also show a qualitative example in Figure~\ref{fig:image-examples}, which highlights the driveable area compliance of trajectories predicted with higher off-road penalties.


\subsubsection{Weighted cross-entropy loss}
\vspace{-2mm}
\begin{table}[tbh]
\centering
\begin{tabular}{lc ccc ccc c}
\toprule
Loss & CE & Avoid Nearby WCE & \multicolumn{3}{c}{Max $\ell^2$ WCE} & \multicolumn{3}{c}{Mean $\ell^2$ WCE} \\
\toprule
Threshold (m) & N/A & 2 & 2 & 3 & 1000 & 2 & 3 & 1000 \\
\midrule
\ade{1} $\downarrow$  & \textbf{1.75} & \textbf{1.75} & 1.79 & 1.84 & 2.01 & 2.05 & 1.91 & 1.94 \\
\ade{5} $\downarrow$  & \textbf{1.05} & \textbf{1.05} & 1.19 & 1.26 & 1.42 & 1.30 & 1.32 & 1.41 \\
\missrate $\downarrow$ & \textbf{0.38} & \textbf{0.38} & 0.44 & 0.49 & 0.55 & 0.48 & 0.49 & 0.53 \\
\midrule
Mean dist. between modes & 1.34 & 1.35 & 1.05 & 1.13 & 1.01 & 1.17 & 1.23 & 0.93 \\
\bottomrule
\end{tabular}
\vspace{2mm}
\caption{Effect of various weighted cross-entropy (WCE) loss functions for CoverNet with $\epsilon = 1m$.}
\label{tab:modified-cross-entropy-loss}
\end{table}

Table~\ref{tab:modified-cross-entropy-loss} compares performance of a dense trajectory set ($\varepsilon=1m$) when we consider all trajectories within various distances of the ground truth as a match.
Surprisingly, the cross-entropy loss performs better than weighted max $\ell^2$ and mean $\ell^2$ cross-entropy losses.
We hypothesize that this is due to the weighted cross-entropy loss causing more ``clumping'' of modes.
By ``clumping'', we mean that the model has difficulty differentiating between nearby trajectories that might have a similar amount of weight associated.
We explored this hypothesis by computing the mean distance between predicted trajectories, and results in Table~\ref{tab:modified-cross-entropy-loss} support this interpretation.
To reduce clumping and still take into account the geometric relationships between trajectories, we experiment with an ``Avoid Nearby'' weighted cross entropy loss that assigns weight of $1$ to the closest match, $0$ to all other trajectories within $2$ meters of ground truth, and $1/|\mathcal{K}|$ to the rest.
We see that we are able to increase mode diversity and recover the performance of the baseline loss.
Our results indicate that losses that are better able to enforce mode diversity may lead to improved performance.

\subsection{Classification vs. ordinal regression}
We also perform a detailed comparison of classification (CoverNet) and ordinal regression (MultiPath), using their standard loss formulations.
All models use the same input representation and backbone to focus on the output representation.

\noindent\textbf{nuScenes}.
\label{subsec:nuscenes}
nuScenes~\cite{nuscenes2019} is a self-driving car dataset that consists of $1,000$ scenes recorded in Boston and Singapore.
Objects are hand-annotated in 3D and published at $2$ Hz.
The task is to predict the next $6$ seconds given the prior $1$ second and the map.
We use the split publicly available in the nuScenes \href{https://www.nuscenes.org/}{software development kit}~\cite{nuScenes-website}, which only includes moving vehicles.
This split includes $32,186$ observations in the train set.

\begin{table}[ht!]
\small
\centering
\begin{tabular}{ll cccc}
\toprule
Method & Modes   & \ade{1} $\downarrow$ & \ade{5} $\downarrow$ & \ade{10} $\downarrow$ & \missrate $\downarrow$ \\
\midrule 
Constant velocity & N/A   & 2.33 $|$ 4.11 & 2.33 $|$ 4.11 & 2.33 $|$ 4.11 & 0.77 $|$ 0.90  \\
Physics oracle &  N/A      & 2.26 $|$ \textbf{2.97} & 2.26 $|$ 2.97 & 2.26 $|$ 2.97 & 0.76 $|$ 0.85 \\
\midrule 
MTP~\cite{cui2019multimodal} & 1 $|$ 1 & 1.80 $|$ 3.41 & 1.80 $|$ 3.41 & 1.80 $|$ 3.41 & 0.71 $|$ 0.90 \\
MultiPath~\cite{chai2019multipath} & 16 $|$ 16 & 1.81 $|$ 4.30 & 1.14 $|$ 2.22 & 1.10 $|$ 2.16 & 0.56 $|$ 0.89 \\
MultiPath~\cite{chai2019multipath} & 64 $|$ 64 & 1.90 $|$ 4.40 & 0.98 $|$ \textbf{1.94} & 0.89 $|$ \textbf{1.68} & 0.41 $|$ 0.84 \\
MultiPath~\cite{chai2019multipath}, $\varepsilon$=3 & 682 $|$ 946 & 1.77 $|$ 4.70 & \textbf{0.91} $|$ 2.27 & \textbf{0.75} $|$ 1.73 & \textbf{0.32} $|$ 0.74 \\
MultiPath~\cite{chai2019multipath}, $\varepsilon$=2 & 1571 $|$ 2339 & 1.76 $|$ 4.80 & 0.94 $|$ 2.50 & 0.76 $|$ 1.85 & \textbf{0.32} $|$ 0.72 \\
\midrule 
CoverNet~\cite{phan2019cover-net}, $\varepsilon$=3 & 682 $|$ 946 & 1.94 $|$ 4.60 & 1.03 $|$ 2.22 & 0.91 $|$ \textbf{1.68} & 0.41 $|$ 0.78 \\
CoverNet~\cite{phan2019cover-net}, $\varepsilon$=2 & 1571 $|$ 2339 & 1.85 $|$ 4.60 & 1.00 $|$ 2.30 & 0.82 $|$ 1.70 & 0.35 $|$ \textbf{0.71} \\
CoverNet~\cite{phan2019cover-net}, $\varepsilon$=1 & 5077 $|$ 9909 & \textbf{1.75} $|$ 6.10 & 1.05 $|$ 2.90 & 0.87 $|$ 2.12 & 0.38 $|$ 0.80 \\
\bottomrule
\end{tabular}
\caption{Results listed as Argoverse $|$ nuScenes, with prediction horizons of $3$ and $6$ sec. Smaller minADE\textsubscript{k} and \missrate is better. CoverNet and MultiPath share trajectory sets. $\varepsilon$ is in meters.}
\label{tab:main-results}
\end{table}
\vspace{-5mm}

\subsubsection{Discussion}

Table~\ref{tab:main-results} shows our comparison between classification and ordinal regression for the CoverNet and MultiPath models using their standard loss formulations. 

For CoverNet models, we find that $\epsilon=2m$ offers the best or second best performance for all metrics on both datasets.
Additionally, smaller $\epsilon$ models tend to overfit, especially on the smaller nuScenes dataset.
For MultiPath, we find that the \missrate monotonically decreases as the trajectory set size increases but using a modest number of modes, such as $64$, can achieve the best \ade{5}.

We emphasize the strong performance of the MultiPath approach with a large number of modes ($\epsilon=3$ and $\epsilon=2$).
The MultiPath paper~\cite{chai2019multipath} noted decreasing performance with more than $64$ anchors, whereas we see benefits from using an order of magnitude more.
Our analysis suggests that the increase in performance associated with the higher number of modes happens due to better coverage of space via anchors, leaving the network to learn smaller residuals.
Table~\ref{tab:residuals} displays the average $\ell^{1}$ and $\ell^{inf}$ norms of the residuals learned by the model using different numbers of modes.

\subsubsection{Argoverse test set}
We further compared our implementation and modifications of MultiPath and CoverNet on the Argoverse test set.
Table~\ref{tab:argoverse-leaderboard} compares our results with notable published methods.
Our multimodal results are similar to MFP~\cite{tang2019mfp}, which explicitly models interactions.
Our poor unimodal performance may be due to the simple input representation we used, which encodes history via fading colors.
\begin{table}[tbh]
\centering
\begin{tabular}{l cccccc}
\toprule
Model & CV~\cite{chang2019argoverse} & LSTM ED~\cite{chang2019argoverse} & VectorNet~\cite{gao2018vector-net} & MFP~\cite{tang2019mfp} & MultiPath\textsuperscript{*}~\cite{chai2019multipath} & CoverNet\textsuperscript{*}~\cite{phan2019cover-net} \\
\midrule
\ade{1} $\downarrow$ & 3.55 & 2.15 & 1.81 & -    & 2.34 & 2.38 \\
\ade{6} $\downarrow$ & 3.55 & 2.15 & 1.18 & 1.40 & 1.28 & 1.42 \\
\bottomrule
\end{tabular}
\vspace{2mm}
\caption{Performance against selected published results on Argoverse test set. CV = constant velocity. MultiPath and CoverNet models (*) use our implementation with a $\varepsilon=2m$ trajectory set.}
\label{tab:argoverse-leaderboard}
\end{table}
\vspace{-5mm}

\section{Conclusion}
We extended a state-of-the-art classification-based motion prediction algorithm to utilize domain knowledge.
By adding an auxiliary loss that penalizes off-road predictions, the model can better learn that likely future motions stay on the road.
Additionally, our formulation makes it easy to \emph{pretrain} using only map information (e.g., off-road area), which significantly improves performance on small datasets.
In an attempt to better encode spatial-temporal relationships among trajectories, we also investigated various weighted cross-entropy losses.
Our results here did not improve on the baseline, although pointed towards the need for losses that promote mode diversity.
Finally, our detailed analysis of classification and ordinal regression (on public self-driving datasets) showed that best performance can be achieved with an order of magnitude more modes than previously reported.


{\small
\bibliographystyle{ieee_fullname}
\bibliography{bibliography}
}

\pagebreak
\section{Supplementary Material}
\label{sec:supplementary}
We created fixed trajectory sets as described in Section 3 of~\cite{phan2019cover-net}.
We used $60,000$ ground truth trajectories from the Argoverse training set and all trajectories from the nuScenes training set to compute trajectory sets.
The parameter $\varepsilon$ controls the furthest a ground truth trajectory in our training set can be from a trajectory in our trajectory set.
We visualize the trajectory sets used by CoverNet and MultiPath on the Argoverse data set and observe that for all $\varepsilon$, the trajectory sets include fast-moving lane keeping trajectories and some very wide turns.
As we decrease $\varepsilon$, the diversity of lateral maneuvers increases.

\begin{figure}[h]
    \centering
    \includegraphics[width=1.0\textwidth]{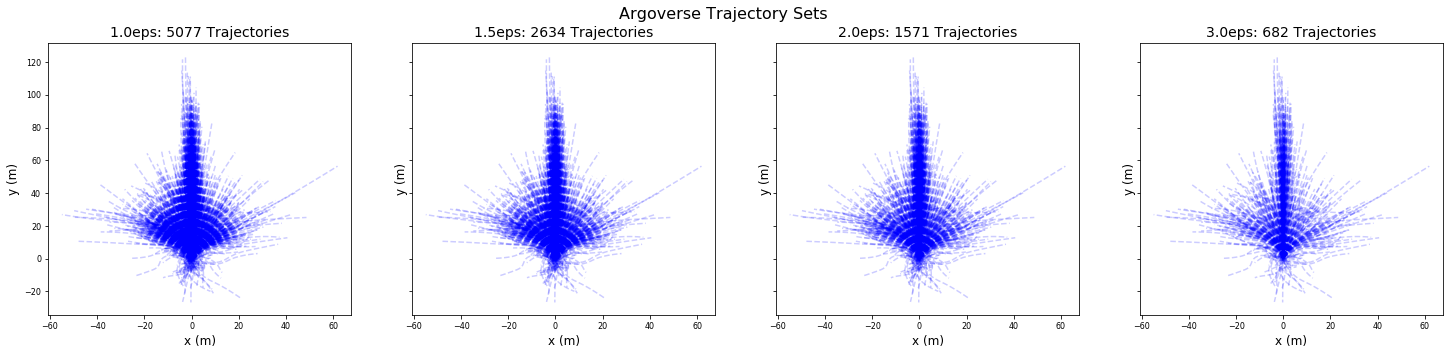}
    \caption{The trajectory sets used by CoverNet and Multipath ($\varepsilon=2, 3$ m) models for Argoverse.}
    \label{fig:traj-set-viz}
\end{figure}


\end{document}